\documentclass[11pt,a4paper]{article}
\usepackage[hyperref]{acl2021}
\usepackage{times}
\usepackage{latexsym}
\usepackage{framed}
\usepackage{amsmath} 
\usepackage{graphicx}
\usepackage{dsfont}
\usepackage{bm}
\usepackage{amsfonts}
\usepackage{listings}
\usepackage{caption}
\usepackage{subcaption}
\usepackage{pythonhighlight}
\usepackage{booktabs}

\usepackage{microtype}

\aclfinalcopy 
\def\aclpaperid{3171} 

\title{Keep it Simple: Unsupervised Simplification of Multi-Paragraph Text}

\author{Philippe Laban \\
  UC Berkeley \\
   \\
  \And
  Tobias Schnabel \\
  Microsoft\\
  \And
  Paul N. Bennett\\
  Microsoft
  \And
  Marti A. Hearst \\
  UC Berkeley\thanks{~Author emails: \{phillab,hearst\}@berkeley.edu, \{Tobias.Schnabel,Paul.N.Bennett\}@microsoft.com} \\
}

\date{}

\begin{document}
\maketitle
\begin{abstract}
    This work presents Keep it Simple (KiS), a new approach to unsupervised text simplification which learns to balance a reward across three properties: fluency, salience and simplicity. We train the model with a novel algorithm to optimize the reward ($k$-SCST), in which the model proposes several candidate simplifications, computes each candidate's reward, and encourages candidates that outperform the mean reward. Finally, we propose a realistic text comprehension task as an evaluation method for text simplification. When tested on the English news domain, the KiS model outperforms strong supervised baselines by more than 4 SARI points, and can help people complete a comprehension task an average of 18\% faster while retaining accuracy, when compared to the original text.
\end{abstract}

\section{Introduction}

The main objective of text simplification is to make a complex text accessible to a wide audience by increasing its readability. In contrast with text summarization -- in which key content is selected to remain in the summary and other content is elided -- in text simplification, ideally all relevant content is preserved.

\begin{figure}[th]
        
        
        
        
    \centering
    \includegraphics[width=0.47\textwidth]{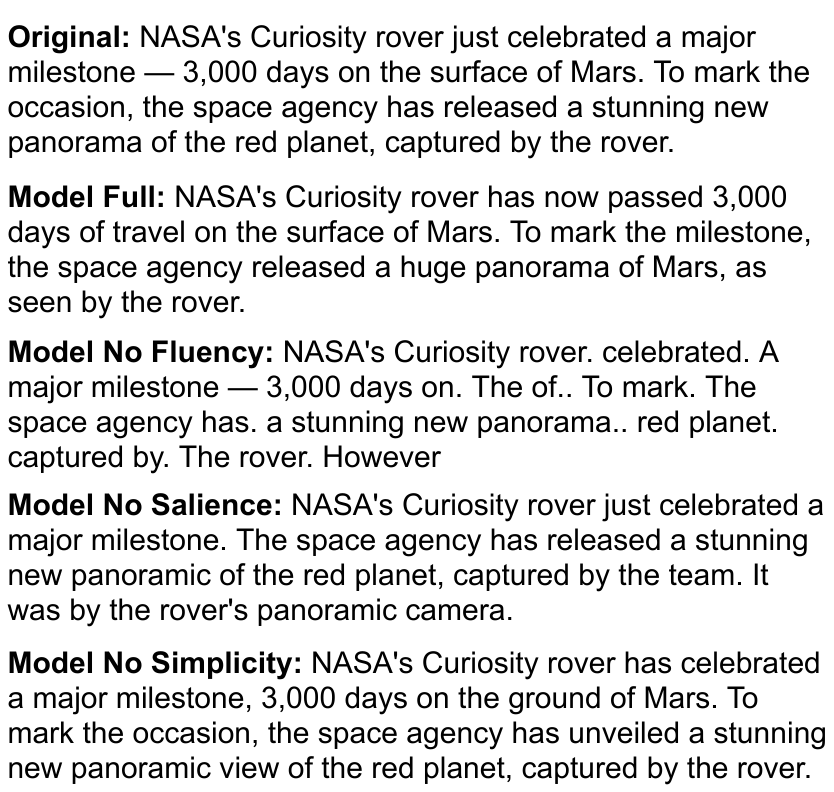}
    \caption{\textbf{Motivating example for the KiS method, based on a CBS article \cite{lewis_2021}.} We optimize a three-component reward: fluency, salience and simplicity. We show model outputs when trained with all three components, and with a missing component.}
    \label{fig:ablation_examples}
\end{figure}

We propose that text simplification algorithms need to balance three properties: (1) \textbf{fluency}: the simplified text should use well-formed English sentences, (2) \textbf{salience}: the simplified text should relay the same information as the original, and (3) \textbf{simplicity}: the simplified text should be syntactically and lexically simpler than the original.

Figure~\ref{fig:ablation_examples} provides intuition for the necessity of each of the three properties. It shows the original text and the output of the full proposed model compared to three reduced versions:

\textbf{Without Fluency}, the generator has no incentive to generate full sentences, and learns it can boost the simplicity score by generating short phrases with excessive punctuation.

\textbf{Without Salience}, the generator does not gain by covering facts in the original text, and  can improve the simplicity score by learning to remove facts (e.g., not mentioning planet Mars by name).

\textbf{Without Simplicity}, the generator is not guided to favor syntactically and lexically simpler re-writes. In Figure \ref{fig:ablation_examples}, \textit{Model No Simplicity} is in fact more complex than the original according to readability measures.

As we show in the related work section (Section~\ref{sec:related_work}), there are no high-quality, large datasets publicly released for text simplification. In this work, we build on recent progress of reinforcement learning (RL)-based training of text generators: we formulate a reference-free reward for text simplification and directly optimize it, circumventing the need for aligned data.

Our main contribution is the Keep it Simple (KiS) procedure, a novel unsupervised method for text simplification. Applied to the English news domain, KiS outperforms several supervised models on common simplification metrics such as SARI \cite{xu2016optimizing} and the Flesch-Kincaid Grade Level \cite{Kincaid1975DerivationON}.

A second contribution is a new algorithm for RL-based training of text generators, $k$-SCST, which is an extension of Self-Critical Sequence Training \cite{rennie2017self}. For each input, we generate $k$ sampled outputs (vs. 2 in SCST), and use the mean population reward as a baseline. We show in Section~\ref{sec:training} that in our domain, $k$-SCST outperforms models trained with SCST.

A third contribution is a novel evaluation method for text simplification. Based on the assumption that simplified text should enable    faster reading  with better understanding, we propose a realistic Text Comprehension task. We show that people reading  texts simplified by KiS are able to complete comprehension tasks faster than comparison texts.

Another departure from previous work is that we work with \emph{paragraphs} as units of text. Most work in text simplification is done at the sentence level, despite work such as \citet{zhong2020discourse} showing that common simplification phenomena occur at the level of the paragraph, (e.g., the deletion, insertion or re-ordering of full sentences). Specifically, we train our models to simplify full paragraphs, and evaluate our models in a human evaluation on short \textit{documents} (i.e., 3-4 paragraphs).

Through rigorous empirical evaluation, we demonstrate the strong performance of our approach;  automated results show that this unsupervised approach is able to outperform strong supervised models by 4 SARI points or more. 
We publicly released the code and model checkpoints\footnote{\url{https://github.com/tingofurro/keep_it_simple}}.

\section{Related Work}

\textbf{Simplification Datasets.} Early datasets were first based on Simple Wikipedia\footnote{\url{https://simple.wikipedia.org/}}: WikiSmall \cite{zhu2010monolingual}, later expanded into WikiLarge \cite{zhang2017sentence}.
\citet{Xu2015ProblemsIC} show there are quality concerns with Simple Wikipedia datasets, and propose Newsela\footnote{\url{https://newsela.com/}} as a replacement. Newsela is a project led by educators re-writing news articles targeting different school grade levels. We view Newsela as the gold-standard for our work, and use the public Newsela release of 1,911 groups of articles to design and evaluate our work. Using a coarse paragraph alignment algorithm, we extract 40,000 paired simple/complex paragraphs targeting a separation of 4 grade levels. We call this dataset the \textit{paired Newsela dataset}, which we use for analysis and baseline training.

\textbf{Seq2Seq for Simplification}. Text simplification is most commonly framed as a sequence-to-sequence (seq2seq) task, leveraging model architectures of other seq2seq tasks, such as natural machine translation \cite{zhu2010monolingual, wubben2012sentence}. \citet{martin2020controllable} introduce ACCESS, a finetuned Transformer model that achieves state-of-the-art performance on WikiLarge. ACCESS can customize simplifications on parameters such as compression rate and paraphrase amount. We directly compare our approach to ACCESS.

Data availability remains one of the main limitations to seq2seq-based text simplification. We side-step this issue entirely by working with unsupervised data, only requiring a small dataset with coarse-level alignments for calibration.


\textbf{Lexical Simplification} focuses on the substitution of single words or phrases with simpler equivalents, with diverse approaches using lexical databases such as WordNet \cite{thomas2012wordnet}, to using contextualized word vectors \cite{qiang2020lexical}. These methods tend to be limited, as they do not consider syntactic complexity, and have no direct way of modeling deletions and insertions. We incorporate a lexical score ($L_{Score}$) as one of the rewards in our simplicity component.

\textbf{Text-edit for Simplification}. Recent work \cite{dong2019editnts, stahlberg2020seq2edits} has modeled text simplification as a \textit{text-edit} task, learning sequences of word-edits that transform the input into the output. Text editing offers explainability, at the cost of added model complexity. We find that without explicitly representing edits, the KiS model easily learns to copy (using attention heads) and deviate from the original text. Outputs can be post-processed into edits, if desired.

\textbf{Unsupervised Simplification} has mostly been limited to lexical simplification. Recently \citet{surya2019unsupervised} (Unsup NTS) proposed a system that can perform both lexical and syntactic simplification, with a joint encoder, and two decoders (simple and complex). We directly compare our unsupervised approach to Unsup NTS.

\textbf{RL for Simplification}. Prior work \cite{zhang2017sentence, guo2018dynamic} used Reinforcement Learning (RL)-based simplification. However, in both cases, components of the reward or training procedure involved reference simplifications, requiring an aligned dataset. By designing a reference-free reward, we are able to train our model with RL without supervision.

\textbf{Evaluation of Simplification}. This usually falls into two categories: automatic offline evaluation, and human evaluation. Automatic evaluations usually involve using n-gram overlap calculations such as BLEU \cite{papineni2002bleu} and SARI \cite{xu2016optimizing}). SARI was shown to correlate better with human judgements of simplicity than BLEU, and it has since become a standard \cite{zhang2017sentence, surya2019unsupervised, martin2020controllable}. In our experiments, we report both SARI and BLEU.

Human evaluation is typically done in an \emph{intrinsic} way -- e.g., by directly rating factors like fluency, simplicity and relevance of model outputs \cite{surya2019unsupervised, wubben2012sentence}. In this work, we propose an extrinsic, task-based protocol. In our comprehension study, we directly measure how much simplified texts can help a human reader answer questions more efficiently. The closest to our evaluation design is that of \citet{angrosh-etal-2014-lexico} with the important difference that we require participants to resubmit after erroneous answers. In pilot studies, we found this step to be crucial for high-quality responses.

\label{sec:related_work}

\section{KiS Components}

\label{sec:keep_it_simple}
\begin{figure}
    \centering
    \includegraphics[width=0.47\textwidth]{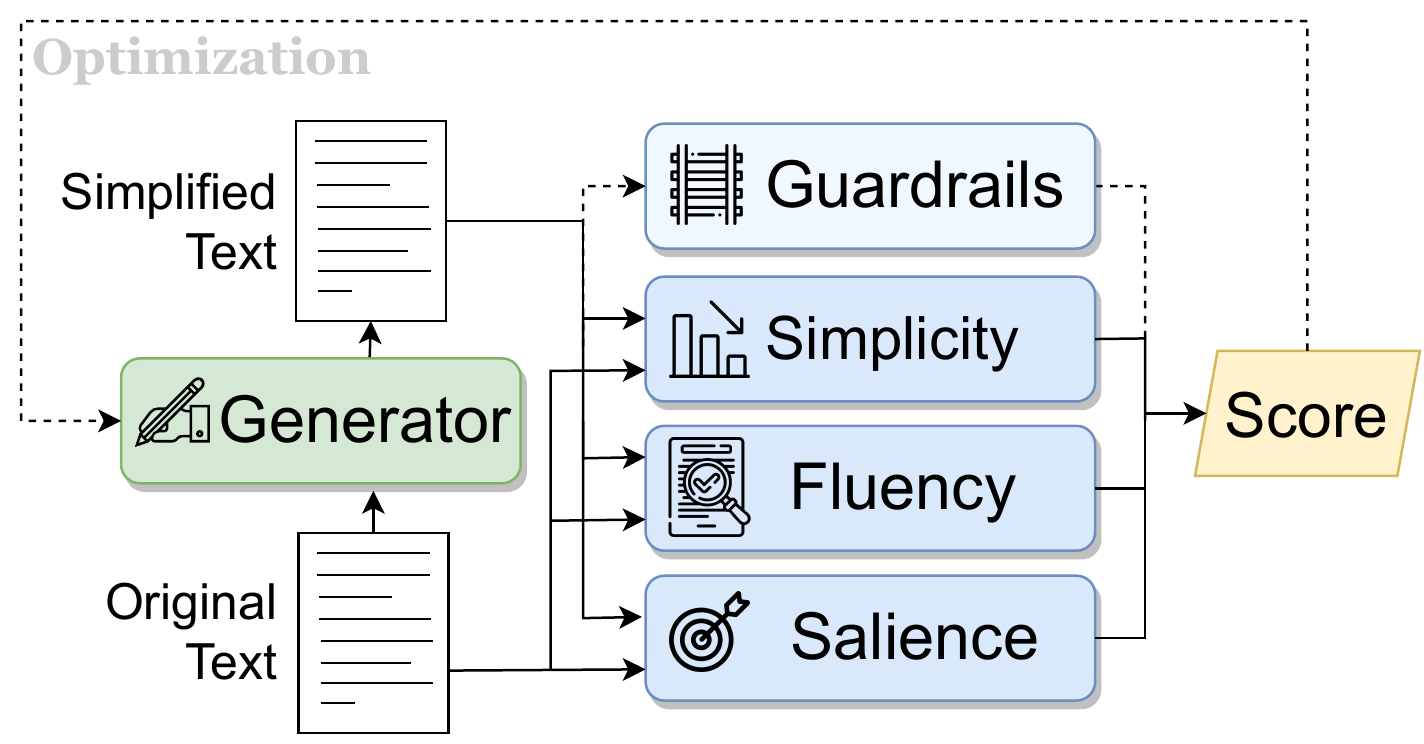}
    \caption{\textbf{Keep it Simple is an unsupervised training procedure for text simplification.} The text generator (GPT-2) produces candidate simplifications, scored according to \textit{fluency}, \textit{simplicity}, \textit{salience}. \textit{Guardrails} enforce the model does not learn high-scoring shortcuts.}
    \label{fig:keep_it_simple_diagram}
\end{figure}

In KiS, we approach unsupervised simplification as a (non-differentiable) reward maximization problem. As shown in Figure~\ref{fig:keep_it_simple_diagram}, there are four components to the reward: simplicity, fluency, salience and guardrails which are jointly optimized. This is essential to avoid trivial solutions that only consider subsets. We therefore use the product of all components as the total reward, because the product is sensitive to the sharp decrease of a single component. For example, the triggering of a single guardrail leads to the zeroing of the total reward. Each component is normalized to the $[0,1]$ range.

\subsection{Simplicity}
\label{sec:simplicity_rewards}

The simplicity score should establish whether the generator's output uses simpler language than the original text. We follow prior work \cite{Ferrs2016YATSYA} and organize our score into a syntactic score $S_{Score}$, and a lexical score $L_{Score}$.  Syntactic simplification focuses on reducing the complexity of a sentence, for example by reducing the number of words in a clause, or reducing distant dependencies. In lexical simplification, the objective is to replace complex phrases with simpler synonyms. To produce a single simplicity score, we take the product of $S_{Score}$ and $L_{Score}$ (both in $[0,1]$).

\subsubsection{Syntactic Simplicity: $\bm{S_{Score}}$}

\begin{figure}
    \centering
    \begin{python}
def S_Score(original,simple):
    Fstart = fkgl(original)
    tgt = target_delta(Fstart)
    Fend   = fkgl(simple)
    D = Fend-Fstart
    return clip(1-((D-tgt)/tgt),0,1)
def target_delta(Fstart):
    # Line-fitted from analysis
    if Fstart < 4.0:
        return 0.1
    if Fstart < 12:
        return 0.5*Fstart-1.9
    return 0.8*Fstart-5.6
    \end{python}
    \caption{\textbf{$\bm{S_{Score}}$ algorithm.} \texttt{fkgl} computes the Flesch-Kincaid grade level.}
    \label{fig:sscore_pseudocode}
\end{figure}

\begin{figure}
    \centering
    \includegraphics[width=0.47\textwidth]{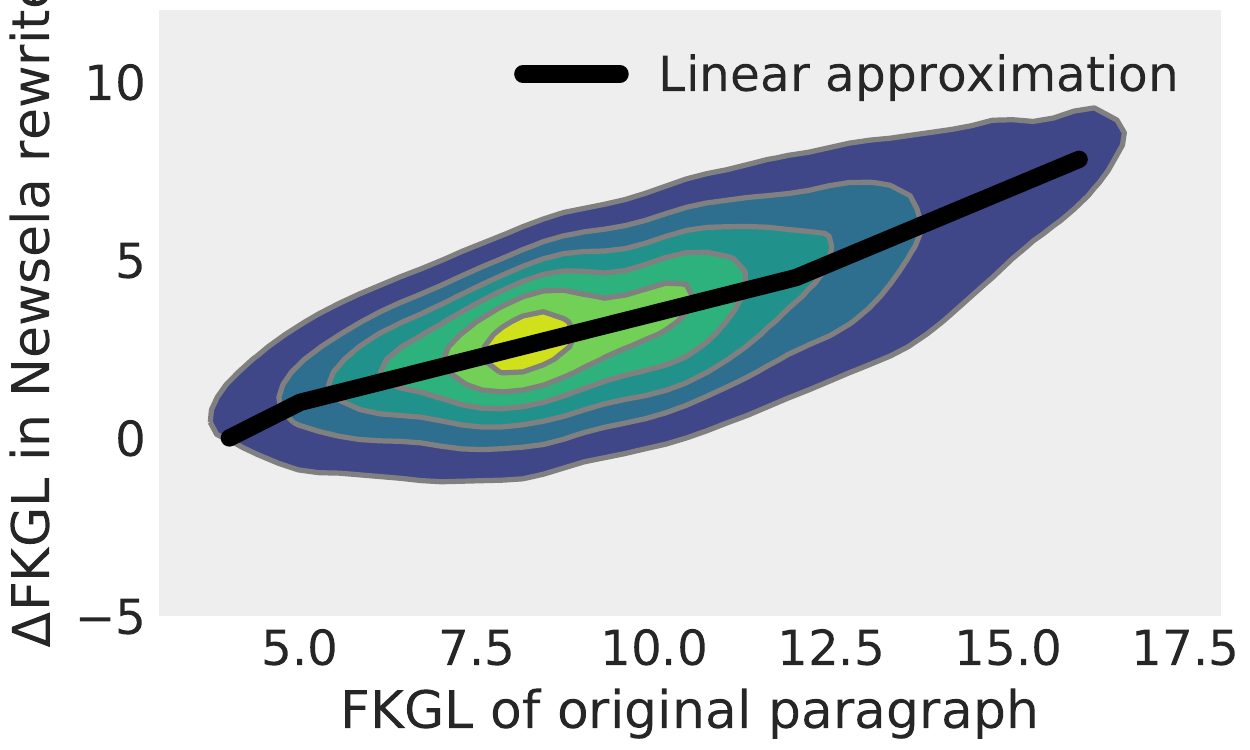}
    \caption{\textbf{Analysis (Kernel Density Estimate plot) of change in Flesch-Kincaid Grade Level in the paired Newsela dataset.} Most simple paragraphs have lower FKGL than the original paragraphs (positive $\Delta FKGL$). When the original paragraph's FKGL is higher (x-axis), the change in FKGL tends to be larger (y-axis). We fit a linear approximation, which we use to compute the $Sscore$.}
    \label{fig:newsela_s_score_analysis}
\end{figure}

We measure syntactic complexity via the Flesch-Kincaid grade level (FKGL) as it is easy to compute and maps to a grade-level which also corresponds to the scale used by Newsela. Other readability metrics such as Dale-Chall formula \cite{dale1948formula}, or the Gunning-Fog index \cite{gunning1969fog} could be used, and future work could examine the effect of choosing one readability metric over the other. Another viable option is the Lexile score \cite{Smith2016TheLF}, however, because its implementation is not publicly released, we cannot use it during training and we report it only for evaluation (done manually on the Lexile Hub\footnote{https://hub.lexile.com}).

Figure~\ref{fig:sscore_pseudocode} shows the $S_{Score}$ algorithm. We compute the original paragraph's FKGL (\verb+FStart+), used to compute a target FKGL (\verb+tgt+). The score is a linear ramp measuring how close the achieved FKGL (\verb+Fend+) is to the target, clipped to $[0,1]$.

In the initial design, the target drop was a constant: 4 grade levels, independent of \verb+FStart+. However, analysis on the paired Newsela corpus revealed that the target FKGL should depend on the initial FKGL. This makes sense intuitively: an already syntactically simple paragraph should not require further simplification, while more complex paragraphs require more simplification. Figure~\ref{fig:newsela_s_score_analysis} shows the positive correlation between the original paragraph's FKGL and the drop of FKGL in the simplified text. We fit a piece-wise linear function to calculate the target FKGL drop from the initial paragraph.

\subsubsection{Lexical Simplicity: $\bm{L_{Score}}$}

Lexical simplicity focuses on whether words in the input paragraph ($W_1$) are more complex than ones in the output paragraph ($W_2$).
We rely on the observation that word frequency and difficulty are correlated \cite{Breland1996WordFA}, and use word frequency in a large corpus of text \cite{Brysbaert2009MovingBK} to determine simplicity.

Because word frequency follows a Zipf power law, we use \citet{robyn_speer_2018_1443582}'s log normalization, adjusting the frequency on a $[0,8]$ range, with words at 0 being non-existent in the corpus, and 8 for most common words. As an example, the word \textit{vigorous} has a frequency of $3.54$, while its more common synonym \textit{strong} obtains $5.23$.

We compute the average Zipf frequency of the set of inserted words ($Z(W_2-W_1)$), and the set of deleted words ($Z(W_1-W_2)$). The difference
\begin{equation}
    \resizebox{0.87\hsize}{!}{%
    $\Delta Z(W_1,W_2) = Z(W_2-W_1) - Z(W_1-W_2)$%
    }
\end{equation}
should be positive.
Analysis of the \textit{paired Newsela corpus} reveals that 91\% of pairs have a positive $\Delta Z(W_1, W_2)$, with a median value of $0.4$. We use this median as the target Zipf shift in the $L_{Score}$, and use a ramp shape similar to the $S_{Score}$, clipped between 0 and 1 (denoted as $[\cdot]^+$):
\begin{equation}
     \resizebox{0.87\hsize}{!}{%
    $L_{Score}(W_1,W_2) = \Bigg[1 - \frac{ \vert \Delta Z(W_1,W_2) - 0.4 \vert }{ 0.4 } \Bigg]^+$%
     }
    \label{eq:SScore}
\end{equation}

\subsection{Fluency}
\label{sec:fluency_rewards}

We use two sub-components for the fluency component: a pre-trained language-model, and a discriminator trained dynamically with the generator.

\subsubsection{Language-Model Fluency}

Language models assign a probability to a sequence of words. This probability is often used to measure fluency of generated text \cite{kann2018sentence, salazar2020masked}. The KiS fluency score is based on a language model in a way similar way to \citet{Laban2020TheSL}. The language model is used to obtain a likelihood of the original paragraph ($LM(p)$) and of the generated output $LM(q)$. We use average log-likelihood, for numerical stability. The language model fluency score is then:
\begin{equation}
    LM_{Score}(p,q) = \Big[1 - \frac{LM(p) - LM(q)}{\lambda}\Big]^+
\end{equation}
$\lambda$ is a tunable hyper-parameter. If the $LM(q)$ is lower than $LM(p)$ by $\lambda$ or more, $LM_{Score}(p,q) = 0$. If $LM(q)$ is above or equal to $LM(p)$, then $LM_{Score}(p,q) = 1$, and otherwise, it is a linear interpolation.

We set $\lambda = 1.3$ as it is the value for which the \textit{paired Newsela dataset} achieves an average $LM_{Score}$ of 0.9.

\subsubsection{Discriminator Fluency}

The $LM_{Score}$ is static and deterministic, which can be limiting, as the generator can learn during training how to adapt and exploit flaws in the language-model (e.g., learning to alter capitalization).

Inspired from the Generative Adversarial Network (GAN) framework \cite{Goodfellow2014GenerativeAN}, we create a dynamic discriminator, trained in conjunction with the generator, dynamically adapting the fluency score during training.

Specifically, we use a RoBERTa model \cite{Liu2019RoBERTaAR} as the basis for the discriminator, a classifier with two labels: 1 for authentic paragraphs, and 0 for generator outputs.

As the generator produces outputs, they are assigned a label of 0 and added to a \textit{training buffer}, while the original paragraphs are assigned a label of 1 and added to the training buffer as well.

Once the training buffer reaches a size of 2,000 samples, the discriminator is trained, using 90\% of the training buffer. We train the discriminator for 5 epochs (details of training are in Appendix~\ref{appendix:training_details}). At the end of each epoch, we checkpoint the discriminator model. We compare the 5 checkpoints in terms of F-1 performance on the remaining 10\% of the training buffer, and keep the best checkpoint as the new discriminator.

The discriminator's probability that a paragraph ($q$) is authentic is the discriminator score:
\begin{equation}
    D_{Score}(q) = p_{disc} ( Y = 1 | X = q )
\end{equation}

As with GANs, there is an equilibrium between the generator attempting to maximize the probability of generating real outputs (``fooling'' the discriminator), and the discriminator succeeding at distinguishing generated and authentic texts.

\subsection{Salience}
\label{sec:salience_reward}

For the salience component, we use the \textit{coverage model} introduced in the summary loop \cite{Laban2020TheSL} for the domain of text summarization, and adapt it to the simplification domain.

The coverage model is a Transformer-based model trained to look at generated text and answer fill-in-the-blank questions about the original text. The score is based on model accuracy at filling in the blanks: the more is filled in, the more relevant the generated content is, and the higher the score.

A key element of the coverage model is its masking procedure, which decides which words to mask. In the summary loop, a limited number of extracted keywords (up to 15 words) are masked. By contrast, for simplification, we mask all non-stop words, amounting to a masking rate of about 40\%.

This change reflects a difference in expectation between summarization and simplification: in summarization, only key components are expected to be recovered from a summary, whereas in simplification most of the original paragraph should be recoverable.
Coverage ranges in $[0,1]$, and reference simplifications in the \textit{paired Newsela corpus} obtain an average score of 0.76, confirming that manual simplification can achieve high coverage.

\subsection{Guardrails}
\label{sec:guardrails}

We use \textit{guardrails} as simple pattern-based scores to avoid common pathological generation problems that we observed. Unlike the main components, guardrails are binary, giving a score of 1 (pass) unless they trigger (score of 0).
We use two guardrails: brevity and inaccuracy.

\subsubsection{Brevity guardrail}

The brevity guardrail ensures the length of generated paragraph ($L_2$) falls in a range around the original paragraph's length ($L_1$). We compute a compression ratio: $C=L_2/L_1$. If $C_{min} \leq C \leq C_{max}$, the guardrail passes, otherwise it triggers.

We set $[C_{min}, C_{max}] = [0.6,1.5]$, because these values ensure the guardrail is not triggered on 98\% of the paired Newsela dataset; this can be adapted depending on the application.

\subsubsection{Inaccuracy guardrail}

Modern text generation models are known to  \textit{hallucinate} facts \cite{huang2020have}, which has led the community to create models to detect and correct hallucinations \cite{cao2020factual, zhang2020optimizing, wang2020asking}.

We propose a light-weight inaccuracy detector  as a guardrail. We use a Named Entity Recognition (NER) model \cite{spacy} to extract entities present in the original paragraph ($E_1$) and the model's output ($E_2$). We trigger the guardrail if an entity present in $E_2$ is not in $E_1$.

Even though human writers can successfully introduce new entities without creating inaccuracies (e.g., replacing the city \textit{La Paz} with the country \textit{Bolivia}), we find that text generators predominantly introduce inaccuracies with novel entities. This simple heuristic can eventually be replaced once inaccuracy detection technology matures.

\section{KiS Training}
\label{sec:training}





\begin{figure}
    \centering
    \includegraphics[width=0.47\textwidth]{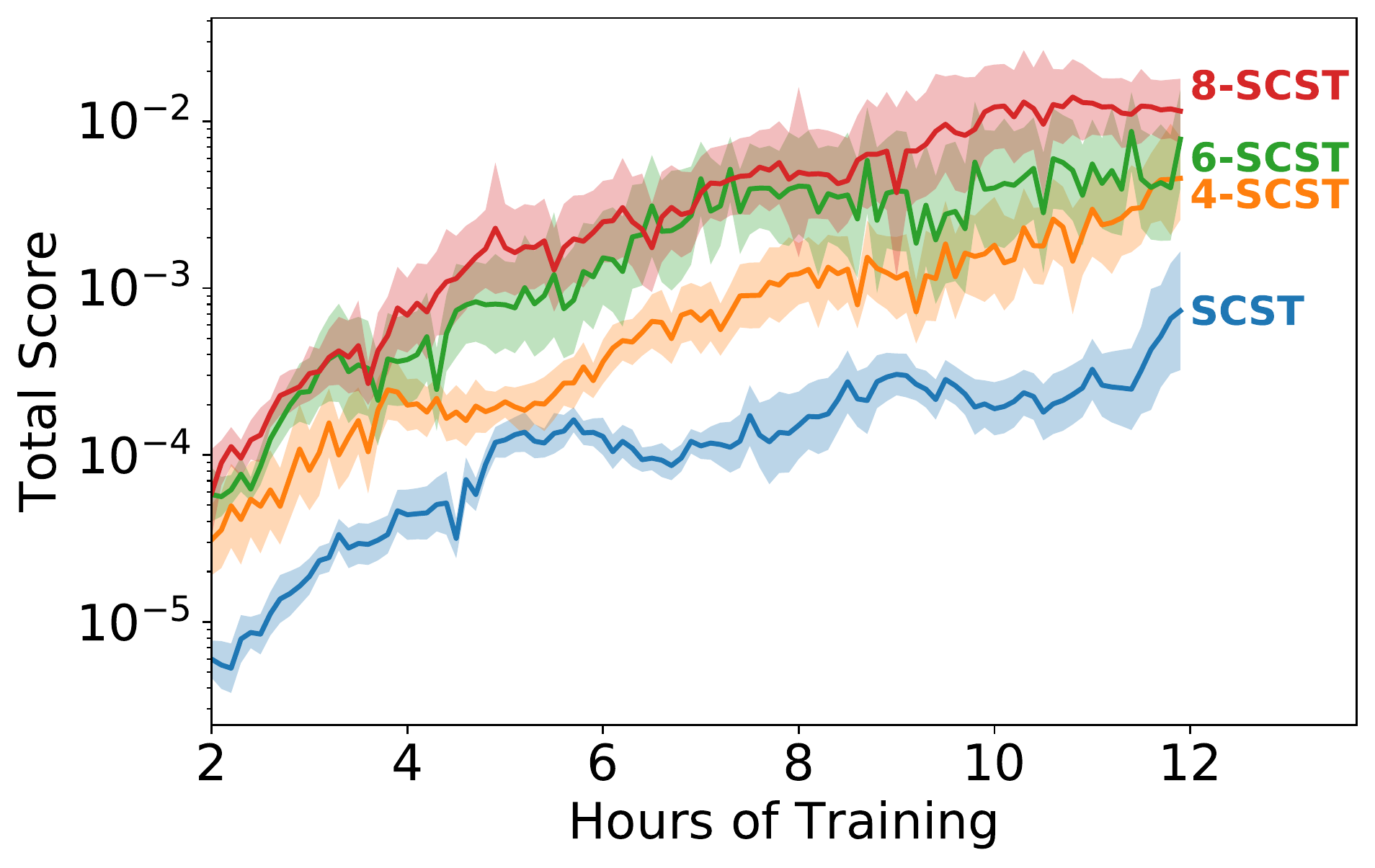}
    \caption{\textbf{Training KiS models comparing SCST with $k$-SCST.} We try 4, 6 and 8 as values for k. Increasing k improves performance and stability.}
    \label{fig:scst_vs_kscst}
\end{figure}

\citet{rennie2017self} introduced Self-Critical Sequence Training (SCST) as an effective algorithm for reward-based training of text generators, successfully applying it to image captioning. The efficacy of SCST was later confirmed on other text generation tasks such as question generation \cite{Zhang2019AddressingSD}, and summarization \cite{celikyilmaz2018deep, Laban2020TheSL}.
In SCST, a probabilistic model is used to generate two distinct candidates: $C^S$, a candidate constructed by sampling the word distribution at each step, and $\hat{C}$, by taking the $\mathrm{argmax}$ of the word distribution at each step. Each candidate is scored, obtaining rewards of $R^S$ and $\hat{R}$, respectively, and the loss is:
\begin{equation}
    L = (\hat{R} - R^S) \sum^N_{i=0} \log p(w_i^S | w_1^S ... w_{i-1}^S, P)
\end{equation}
\noindent
where $p(w_i^S | ...)$ represents the probability of the $i$-th word conditioned on previously generated sampled sequence according to the model, P is the input paragraph, and N the number of words in the generated sequence.
Intuitively, minimizing this loss increases the likelihood of the sampled sequence if $R^S > \hat{R}$, and decreases it otherwise, both increasing the expected total reward.

One limitation in SCST occurs when the two sequences achieve comparable rewards ($R^S \simeq \hat{R}$): the loss nears zero, and the model has little to learn, wasting a training sample. In our experiments with SCST, this can occur with 30\% of samples.

We propose an extension of SCST, which we call $k$-SCST. We generate $k$ sampled candidates ($k>2$), compute the rewards of each candidate $R^{S1}, ..., R^{Sk}$, as well as the mean reward achieved by this sampled population: $\bar{R^S}=(R^{S1}+...+R^{Sk}) / k$, which we use as the baseline, instead of $\hat{R}$. The loss $L$ becomes:
\begin{equation}
    L=\sum^{k}_{j=1} (\bar{R^S} - R^{Sj}) \sum^N_{i=0} \log p(w_i^{Sj} | w_1^{Sj} ... w_{i-1}^{Sj}, P)
\end{equation}

We use a GPT2-medium for the generator, initialized with the released pre-trained checkpoint. Experimental details such as data and optimizer used are provided in Appendix~\ref{appendix:training_details}.

In Figure~\ref{fig:scst_vs_kscst}, we show results of a direct comparison of SCST ($k=2$) with $k$-SCST varying $k$ in $\{4, 6, 8\}$, while keeping other components of the training fixed. Because of the variance involved in RL training, we recorded six independent training runs for each setting (for a total of 24 runs), and plot the average reward across runs of a setting, as well as the standard error of the mean (SEM).

We observe that increasing $k$ leads to higher average reward, and less variation in the reward. In our setting, $k$-SCST boosts performance and stabilizes training. We use $k=8$ in all final models, as increasing $k$ further is impractical due to GPU memory limitations.

We believe $k$-SCST's advantage stems from two factors: first, obtaining a better estimate of the distribution of rewards by sampling more outputs, second, by using the mean reward as the baseline, saving on computation of a separate baseline generation. We believe $k$-SCST can also improve learning in other text generation applications and plan to pursue this in future work.

\section{Experiments}
\label{sec:results}

We present results experimentally validating the KiS procedure for text simplification.
We give results based on automatic metrics, on a novel human comprehension task, and from an ablation study.

\subsection{Models Compared}
  
We compare the \textbf{KiS Model} to three strong supervised models, and an unsupervised approach.

\textbf{ACCESS} from \cite{martin2020controllable}, is a state-of-the-art Transformer model trained on WikiLarge (300,000 pairs of complex/simple sentences). This model uses default parameters ($\mathit{NBChar}$=0.95, $\mathit{LevSim}$=0.75).

\textbf{ACCESS90} is identical to \textbf{ACCESS}, with different parameters ($\mathit{NBChar}$=0.90, $\mathit{LevSim}$=0.75), reducing target compression from 95\% to 90\%, matching the average compression rate in Newsela.

\textbf{Finetune Baseline} is a GPT2-medium model finetuned on the \textit{paired Newsela dataset}. Large pre-trained models often perform competitively in low-resource environments, making this a strong point of comparison.

\textbf{Unsup NTS} from \cite{surya2019unsupervised} is an unsupervised approach based on successively encoding and denoising text using a GRU architecture.

Training details for the KiS Model and Finetune Baseline are in Appendix~\ref{appendix:training_details}.

\subsection{Automatic Results}
  
\begin{table}[]
    \resizebox{0.47\textwidth}{!}{%
    \begin{tabular}{lrrrrrr}
    \toprule
    Model &      SARI &      BLEU & \%FKGL & \%Lexile &  Comp. &  Cov. \\
    \midrule 
    Newsela             &  - &  - & 87 & \textbf{79} & .918 &  .754 \\ \hline

    Finetune Baseline &  .470 &  \textbf{.719} & 68 & 52 & .903 &  \textbf{.894} \\
    ACCESS Default    &  .666 &  .649 & 86 & 63 & .958 &  .805 \\
    ACCESS 90         &  .674 &  .644 & 93 & 64 & .921 &  .789 \\ \hline
    Unsup NTS         &  .677 & .535 & 48 & 57 & .753 &  .618 \\
    KiS Model         &  \textbf{.709} &  .526 & \textbf{100} & 72 & .852 &  .640 \\
    \bottomrule
    \end{tabular}
    }
    \caption{\textbf{Automatic results on Newsela test-set.} \textit{SARI} and \textit{BLEU} are reference-based metrics. \textit{\%FKGL} and \textit{\%Lexile} are percentages of model outputs lowering the grade level. \textit{Comp.} is the average compression ratio (\# words), and \textit{Cov.} the output's average coverage score.}
    \label{tab:automatic_results}
\end{table}
  
We put aside 500 samples from the \textit{paired Newsela dataset} as a test set to compare models on automatic metrics. We compare models on SARI and BLEU, report the percentage when readability measures see an improvement in readability: \%FKGL, and \%Lexile and compute the average compression rate (Comp.), and coverage (Cov.). Results are summarized in Table~\ref{tab:automatic_results}.

The KiS model achieves the highest SARI score by a margin of 0.04, even though it is an unsupervised approach. 

Finetune Baseline achieves the highest BLEU and salience scores, but lowest SARI score. We interpret this as showing the model takes the least risk: high salience, with little simplification.

We observe that all models are able to increase readability in terms of FKGL and Lexile compared to original paragraphs. We note that for almost all models, the percentage is lower for the Lexile measure than for FKGL, showing that an improvement in Lexile score is more difficult to achieve than FKGL. The KiS model achieves an increase in Lexile readability 72\% of the time, the closest figure to 79\% of the Newsela human-written reference.

We note that the perfect performance of KiS on \%FKGL could be explained by the fact that FKGL is a part of a component being optimized ($S_{Score}$), however Lexile was not.

In terms of compression, the KiS model compresses the second most, most likely hurting its coverage. Adjusting the Brevity guardrail could encourage the model to compress less. ACCESS90 has the compression rate closest to Newsela references, but this only leads to a modest improvement in SARI when compared to ACCESS.

Overall, the Newsela references achieve the best percentage of Lexile readability improvement, while outperforming the KiS model at coverage: there is still a gap between human-written simplifications and model-generated ones.

\subsection{Human Comprehension Study}

\begin{figure*}
    \centering
    \includegraphics[width=0.95\textwidth]{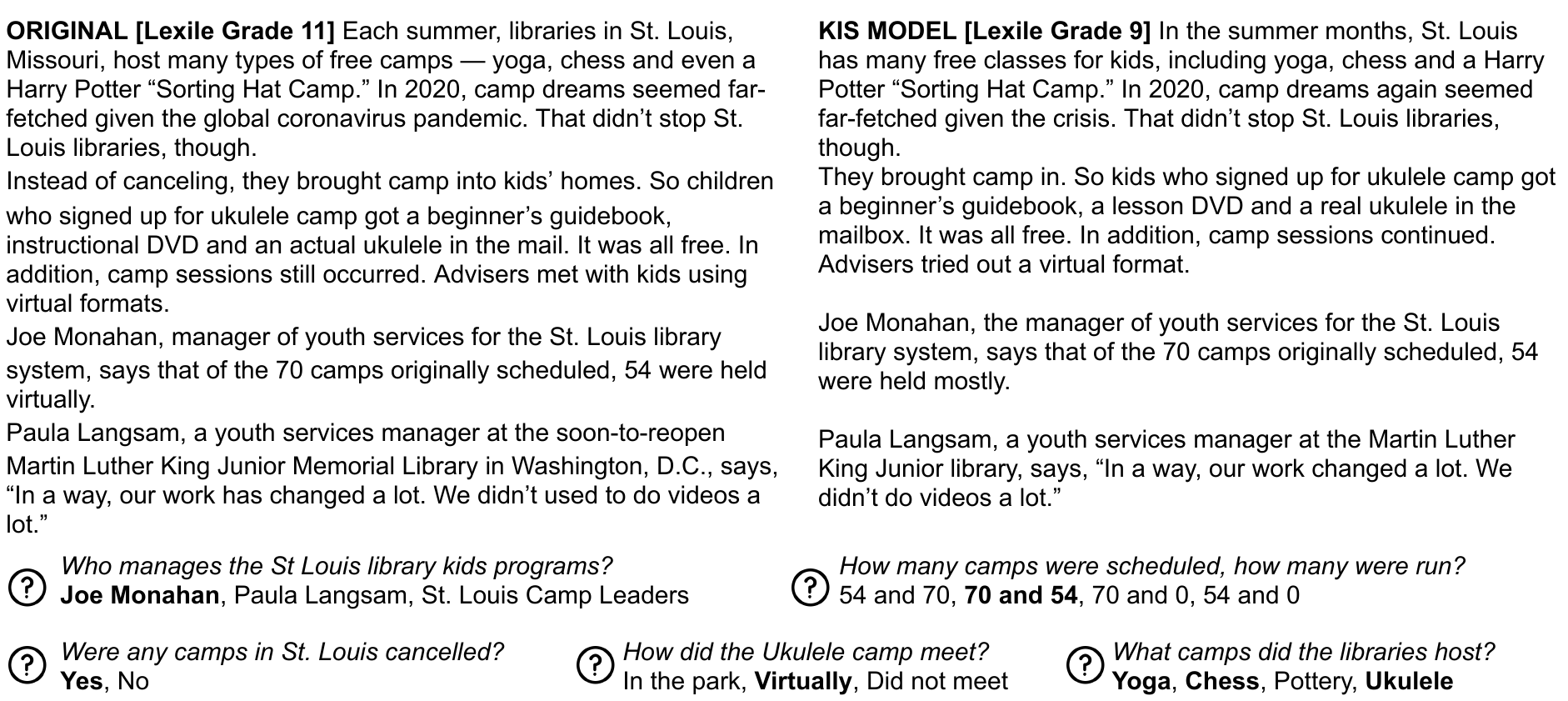}
    \caption{\textbf{Example Task (from a Washington Post article \cite{kelati_2020}) for the Comprehension Study.} Shown are two of five stimuli: original document (left), and KiS model output (right). Participants read a text and answered comprehension questions (bottom). Average completion time was 160 seconds (original) and 136 seconds (KiS model output).}
    \label{fig:user_study_example}
\end{figure*}

We propose a human comprehension study to evaluate the usefulness of simplification results. Simplified text should be easier to read than the original text, while retaining accuracy and understanding. We design a task to evaluate how well both manual and automated simplifications achieve this objective.  The main idea is to show readers a text and ask them to answer multiple-choice questions, evaluating the texts based on time and retries  needed to select the correct answer.

\subsubsection{Study Design}
Five different versions of each document were generated as stimuli: the original document, the Newsela reference, and versions from the three best-performing methods from the last section: KiS, Finetune Baseline, and ACCESS. We did not include Unsup NTS in our analysis, because of its low performance on \%FKGL and \%Lexile metrics. Associated with each document are five manually generated multiple-choice questions, each with one or more correct answers and one to four distractors. The original and the Newsela texts were checked manually by experimenters to ensure that all allow for questions to be answered correctly.
Crowd-workers were shown four documents in succession, in a between-participants design. Order of document and stimuli type were randomized.
Figure~\ref{fig:user_study_example} shows two stimuli of a document (original and KiS) along with the comprehension questions. (The entire set of five stimuli can be found in Figure~\ref{fig:complement_study_examples} in the Appendix.) 

After several rounds of pilot testing, we arrived at the following design choices:

\textbf{Document theme.} We chose recent news articles involving complex themes (e.g., trajectory of iceberg) as the source of documents. For news articles, recency seems to engage participants, and technical terms increase the impact of simplification.

\textbf{Section length.} We chose document length of 3-4 paragraphs (or 200 words), and five comprehension questions. Document length should not be too W (makes some questions trivial), or too long (adds a retrieval component to the task).

\textbf{Selection of questions.} Questions were generated via a GPT2 question generation model finetuned on the NewsQA dataset \cite{trischler2017newsqa}. We select questions answerable by both the original and Newsela references, attempting to have both factoid (answer is entity) and reasoning questions.

\textbf{Re-submission until correct.} When submitting answers, participants received feedback on which were incorrect, and were required to re-submit until all answers were correct. This aligns the objective of the participant (i.e., finishing the task rapidly), with the task's objective (i.e., measuring participant's efficiency at understanding). This also gives a way to discourage participants from ``brute-forcing'' the task, re-submitting many combinations until one works.

We note that some components of the study such as the choice of document themes and the selection of comprehension questions are elements that create variability in the results. We release the models used in the study, as well all generated texts that were evaluated to enable follow-up research and to aid reproducibility.

\subsubsection{Study Results}

We ran the study on Mechanical Turk, accepting crowd-workers with 1700+ completed tasks, and an acceptance rate of 97\%+. The study was active for two weeks in December 2020, and remunerated participants completing all four sections at a rate of \$10/hour. (Appendix~\ref{appendix:user_instructions} shows crowd-worker instructions and the document/version distributions.) When removing ``brute-forced'' submissions (10+ re-submissions), we are left with 244 submissions, used for result analysis reported in Table~\ref{tab:study_results}, (A more detailed results table is included in Appendix~\ref{appendix:full_study_table}.)

We measure two outcomes: question completion time (in seconds), and number of submissions to correctness. We performed a Kruskal-Wallis test \cite{kruskal1952use} with a Dunn post-hoc test \cite{dunn1964multiple} for statistical significance between pairs of conditions.

\begin{table}[]
    \centering
    \resizebox{0.47\textwidth}{!}{%
    \begin{tabular}{lllcc}
    \toprule
    Model                 & Time (sec)          & \# Subs. & Comp. & CASpeed \\
    \midrule
    $\flat$ Original              & 174.0  & 4.23 & 1.0  & 1.00 \\
    $\natural$ Newsela               & 163.3 & 5.10 & 1.08 & \textbf{1.15} \\
    \hline
    $\forall$ ACCESS                & 188.5  & 6.69 & 0.96 & 0.88 \\
    $\exists$ Finetune Baseline   & 161.0 $\forall$ & 4.70 & 0.97 & 1.04 \\
    $\nabla$ KiS Model             & \textbf{142.6} $\flat$ $\natural$ $\forall$ & 4.10 $\forall$ & 0.87 & \textbf{1.06} \\
    \bottomrule
    \end{tabular}
    }
    \caption{\textbf{Results of the Human Comprehension Study.} We measure average completion time (Time), number of submissions (\#Subs.), compression ratio (Comp.) and a compression-accounted speed-up (CASpeed). Each text version is assigned a symbol used to indicate statistical significance ($p<0.05$).}
    \label{tab:study_results}
\end{table}

In line with study objectives, simplified texts help participants complete the task faster than reading original texts, with three of the four simplified versions leading to improvements in completion times. Participants were fastest with KiS simplifications (18\% faster).
The KiS model led to a statistically significant speed-up compared to the originals, Newsela references, and ACCESS simplifications. ACCESS simplifications surprisingly led to a non-significant slow-down, which we attribute to a potential loss in fluency that might have confused participants.

One important factor we consider is that shorter passages (i.e., smaller compression) might lead to a speed-up regardless of simplicity. We confirm this by finding a small positive correlation between passage length and completion time of 0.09. We compute a \textit{compression-adjusted speed-up} (\textit{CASpeed}) ratio by: (1) computing the passage length of each simplified version, (2) linearly extrapolating the expected completion time for this passage length for original paragraphs, and (3) computing the ratio of the extrapolation to the observed completion time. If $CASpeed > 1$, participants were faster than expected for the passage length. Newsela reference paragraphs achieve the best \textit{CASpeed}, followed by the KiS model. This suggests that good simplification can involve making texts longer.

\subsection{Ablation Study}

We train three ablated models, each missing a reward component to gain understanding in the value of each component of the KiS procedure.

Figure~\ref{fig:ablation_examples} gives a qualitative perspective on each ablation. Without fluency, the generator learns to generate incomplete sentences, without salience, it omits important information, and without simplicity, it can sometimes ``complexify''.

We computed complete automatic results for the ablated models, and find that each ablation leads to a decrease on an evaluation metric, confirming that all three components are necessary to generate high-quality simplifications (details in  Appendix~\ref{appendix:ablation_study}).

\label{sec:ablation_study}

\section{Limitations and Future Work}

\textbf{Improved Accuracy Scoring}. The current  guardrail for inaccuracy is rudimentary;  trained models still generate non-factual simplifications. Recent work in fact-checking for the summarization domain \cite{kryscinski2020evaluating, li2018ensure} could be adapted to the simplification domain to improve this.

\textbf{Inclusion of Supervised Signal}. In this work, we establish that text simplification can be approached in an unsupervised manner. In future work, Keep it Simple could be used as a pre-training strategy, or used jointly with supervised training.

\textbf{Reproducibility of Human Evaluation}. Even though we release the models, stimuli and comprehension questions used in the human evaluation, some elements of the procedure introduce randomness. Participating crowd-workers differ in literacy level which may have an effect on their performance at the task \cite{alonzo2021comparison}.

\textbf{New Settings, Domains and Languages}. We limited our experiments to the simplification of English news articles following prior work, but plan to pursue other languages in the future. Similarly, because Keep it Simple does not require labeled data, it can be applied to new settings (e.g., rewriting to inverse the effects of simplification), or to new domains (e.g., legal texts).

\section{Conclusion}

We have shown that text simplification can be approached in an unsupervised manner via KiS. By optimizing a reward comprised of simplicity, fluency and salience components, KiS is able to outperform strong supervised models on automatic metrics (+0.04 in SARI).
We propose a human comprehension task to evaluate the usefulness of simplification and show that simplifications tend to lead to a measurable speed-up in task completion, with KiS texts producing the best speed-up of 18\% on average.
These are first steps for unsupervised text simplification, and we suggest that future work should focus on adapting the methodology to new domains (i.e., legal), non-English languages, and refining optimized rewards to take factuality into account.

\section*{Acknowledgments}

We would like to thank Katie Stasaski, Dongyeop Kang, and the ACL reviewers for their helpful comments, as well as Newsela for providing a version of their simplified news corpus. This work was supported by a Microsoft BAIR Commons grant as well as a Microsoft Azure Sponsorship.

\bibliography{anthology,acl2021}

\begin{thebibliography}{44}
\expandafter\ifx\csname natexlab\endcsname\relax\def\natexlab#1{#1}\fi

\bibitem[{Alonzo et~al.(2021)Alonzo, Trussell, Dingman, and
  Huenerfauth}]{alonzo2021comparison}
Oliver Alonzo, Jessica Trussell, Becca Dingman, and Matt Huenerfauth. 2021.
\newblock Comparison of methods for evaluating complexity of simplified texts
  among deaf and hard-of-hearing adults at different literacy levels.
\newblock In \emph{Proceedings of the 2021 CHI Conference on Human Factors in
  Computing Systems}, pages 1--12.

\bibitem[{Angrosh et~al.(2014)Angrosh, Nomoto, and
  Siddharthan}]{angrosh-etal-2014-lexico}
Mandya Angrosh, Tadashi Nomoto, and Advaith Siddharthan. 2014.
\newblock \href {https://www.aclweb.org/anthology/C14-1188} {Lexico-syntactic
  text simplification and compression with typed dependencies}.
\newblock In \emph{Proceedings of {COLING} 2014, the 25th International
  Conference on Computational Linguistics: Technical Papers}, pages 1996--2006,
  Dublin, Ireland. Dublin City University and Association for Computational
  Linguistics.

\bibitem[{Breland(1996)}]{Breland1996WordFA}
H.~Breland. 1996.
\newblock Word frequency and word difficulty: A comparison of counts in four
  corpora.
\newblock \emph{Psychological Science}, 7:96 -- 99.

\bibitem[{Brysbaert and New(2009)}]{Brysbaert2009MovingBK}
M.~Brysbaert and B.~New. 2009.
\newblock Moving beyond kucera and francis: A critical evaluation of current
  word frequency norms and the introduction of a new and improved word
  frequency measure for american english.
\newblock \emph{Behavior Research Methods}, 41:977--990.

\bibitem[{Cao et~al.(2020)Cao, Dong, Wu, and Cheung}]{cao2020factual}
Meng Cao, Yue Dong, Jiapeng Wu, and Jackie Chi~Kit Cheung. 2020.
\newblock Factual error correction for abstractive summarization models.
\newblock In \emph{Proceedings of the 2020 Conference on Empirical Methods in
  Natural Language Processing (EMNLP)}, pages 6251--6258.

\bibitem[{Celikyilmaz et~al.(2018)Celikyilmaz, Bosselut, He, and
  Choi}]{celikyilmaz2018deep}
Asli Celikyilmaz, Antoine Bosselut, Xiaodong He, and Yejin Choi. 2018.
\newblock Deep communicating agents for abstractive summarization.
\newblock In \emph{Proceedings of the 2018 Conference of the North American
  Chapter of the Association for Computational Linguistics: Human Language
  Technologies, Volume 1 (Long Papers)}, pages 1662--1675.

\bibitem[{Dale and Chall(1948)}]{dale1948formula}
Edgar Dale and Jeanne~S Chall. 1948.
\newblock A formula for predicting readability: Instructions.
\newblock \emph{Educational research bulletin}, pages 37--54.

\bibitem[{Dong et~al.(2019)Dong, Li, Rezagholizadeh, and
  Cheung}]{dong2019editnts}
Yue Dong, Zichao Li, Mehdi Rezagholizadeh, and Jackie Chi~Kit Cheung. 2019.
\newblock Editnts: An neural programmer-interpreter model for sentence
  simplification through explicit editing.
\newblock In \emph{Proceedings of the 57th Annual Meeting of the Association
  for Computational Linguistics}, pages 3393--3402.

\bibitem[{Dunn(1964)}]{dunn1964multiple}
Olive~Jean Dunn. 1964.
\newblock Multiple comparisons using rank sums.
\newblock \emph{Technometrics}, 6(3):241--252.

\bibitem[{Ferr{\'e}s et~al.(2016)Ferr{\'e}s, Marimon, Saggion
  et~al.}]{Ferrs2016YATSYA}
Daniel Ferr{\'e}s, Montserrat Marimon, Horacio Saggion, et~al. 2016.
\newblock Yats: yet another text simplifier.
\newblock In \emph{International Conference on Applications of Natural Language
  to Information Systems}, pages 335--342. Springer.

\bibitem[{Goodfellow et~al.(2014)Goodfellow, Pouget-Abadie, Mirza, Xu,
  Warde-Farley, Ozair, Courville, and Bengio}]{Goodfellow2014GenerativeAN}
Ian Goodfellow, Jean Pouget-Abadie, Mehdi Mirza, Bing Xu, David Warde-Farley,
  Sherjil Ozair, Aaron Courville, and Yoshua Bengio. 2014.
\newblock \href
  {https://proceedings.neurips.cc/paper/2014/file/5ca3e9b122f61f8f06494c97b1afccf3-Paper.pdf}
  {Generative adversarial nets}.
\newblock In \emph{Advances in Neural Information Processing Systems},
  volume~27.

\bibitem[{Gunning(1969)}]{gunning1969fog}
Robert Gunning. 1969.
\newblock The fog index after twenty years.
\newblock \emph{Journal of Business Communication}, 6(2):3--13.

\bibitem[{Guo et~al.(2018)Guo, Pasunuru, and Bansal}]{guo2018dynamic}
Han Guo, Ramakanth Pasunuru, and Mohit Bansal. 2018.
\newblock Dynamic multi-level multi-task learning for sentence simplification.
\newblock In \emph{Proceedings of the 27th International Conference on
  Computational Linguistics}, pages 462--476.

\bibitem[{Honnibal et~al.(2020)Honnibal, Montani, Van~Landeghem, and
  Boyd}]{spacy}
Matthew Honnibal, Ines Montani, Sofie Van~Landeghem, and Adriane Boyd. 2020.
\newblock \href {https://doi.org/10.5281/zenodo.1212303} {{spaCy:
  Industrial-strength Natural Language Processing in Python}}.
\newblock Doi.org/10.5281/zenodo.1212303.

\bibitem[{Huang et~al.(2020)Huang, Cui, Yang, Bao, Wang, Xie, and
  Zhang}]{huang2020have}
Dandan Huang, Leyang Cui, Sen Yang, Guangsheng Bao, Kun Wang, Jun Xie, and Yue
  Zhang. 2020.
\newblock What have we achieved on text summarization?
\newblock In \emph{Proceedings of the 2020 Conference on Empirical Methods in
  Natural Language Processing (EMNLP)}, pages 446--469.

\bibitem[{Kann et~al.(2018)Kann, Rothe, and Filippova}]{kann2018sentence}
Katharina Kann, Sascha Rothe, and Katja Filippova. 2018.
\newblock Sentence-level fluency evaluation: References help, but can be
  spared!
\newblock In \emph{Proceedings of the 22nd Conference on Computational Natural
  Language Learning}, pages 313--323.

\bibitem[{Kelati(2020)}]{kelati_2020}
Haben Kelati. 2020.
\newblock \href
  {https://www.washingtonpost.com/lifestyle/kidspost/librarians-find-creative-ways-to-serve-kids-when-buildings-are-closed-for-browsing/2020/09/22/fd6f2db4-f9b6-11ea-a275-1a2c2d36e1f1_story.html}
  {Librarians find creative ways to serve kids when buildings are closed for
  browsing}.
\newblock \emph{The Washington Post}.

\bibitem[{Kincaid et~al.(1975)Kincaid, Fishburne~Jr., Rogers, and
  Chissom}]{Kincaid1975DerivationON}
J.~Peter Kincaid, Robert~P. Fishburne~Jr., Richard~L. Rogers, and Brad~S.
  Chissom. 1975.
\newblock Derivation of new readability formulas (automated readability index,
  fog count and flesch reading ease formula) for navy enlisted personnel.
\newblock Technical report, Naval Technical Training Command Millington TN
  Research Branch.

\bibitem[{Kruskal and Wallis(1952)}]{kruskal1952use}
William~H Kruskal and W~Allen Wallis. 1952.
\newblock Use of ranks in one-criterion variance analysis.
\newblock \emph{Journal of the American statistical Association},
  47(260):583--621.

\bibitem[{Kryscinski et~al.(2020)Kryscinski, McCann, Xiong, and
  Socher}]{kryscinski2020evaluating}
Wojciech Kryscinski, Bryan McCann, Caiming Xiong, and Richard Socher. 2020.
\newblock Evaluating the factual consistency of abstractive text summarization.
\newblock In \emph{Proceedings of the 2020 Conference on Empirical Methods in
  Natural Language Processing (EMNLP)}, pages 9332--9346.

\bibitem[{Laban et~al.(2020)Laban, Hsi, Canny, and Hearst}]{Laban2020TheSL}
Philippe Laban, Andrew Hsi, John Canny, and Marti~A. Hearst. 2020.
\newblock \href {https://doi.org/10.18653/v1/2020.acl-main.460} {The summary
  loop: Learning to write abstractive summaries without examples}.
\newblock In \emph{Proceedings of the 58th Annual Meeting of the Association
  for Computational Linguistics}, pages 5135--5150. Association for
  Computational Linguistics.

\bibitem[{Lewis(2021)}]{lewis_2021}
Sophie Lewis. 2021.
\newblock \href
  {https://www.cbsnews.com/news/nasa-curiosity-rover-celebrates-3000-day-mars-panorama/}
  {Nasa curiosity rover celebrates 3,000th day on mars with stunning panorama
  of planet}.
\newblock \emph{CBS News}.

\bibitem[{Li et~al.(2018)Li, Zhu, Zhang, and Zong}]{li2018ensure}
Haoran Li, Junnan Zhu, Jiajun Zhang, and Chengqing Zong. 2018.
\newblock Ensure the correctness of the summary: Incorporate entailment
  knowledge into abstractive sentence summarization.
\newblock In \emph{Proceedings of the 27th International Conference on
  Computational Linguistics}, pages 1430--1441.

\bibitem[{Liu et~al.(2019)Liu, Ott, Goyal, Du, Joshi, Chen, Levy, Lewis,
  Zettlemoyer, and Stoyanov}]{Liu2019RoBERTaAR}
Y.~Liu, Myle Ott, Naman Goyal, Jingfei Du, Mandar Joshi, Danqi Chen, Omer Levy,
  M.~Lewis, Luke Zettlemoyer, and Veselin Stoyanov. 2019.
\newblock Roberta: A robustly optimized bert pretraining approach.
\newblock \emph{ArXiv}, abs/1907.11692.

\bibitem[{Martin et~al.(2020)Martin, de~la Clergerie, Sagot, and
  Bordes}]{martin2020controllable}
Louis Martin, {\'E}ric~Villemonte de~la Clergerie, Beno{\^\i}t Sagot, and
  Antoine Bordes. 2020.
\newblock Controllable sentence simplification.
\newblock In \emph{Proceedings of The 12th Language Resources and Evaluation
  Conference}, pages 4689--4698.

\bibitem[{Papineni et~al.(2002)Papineni, Roukos, Ward, and
  Zhu}]{papineni2002bleu}
Kishore Papineni, Salim Roukos, Todd Ward, and Wei-Jing Zhu. 2002.
\newblock Bleu: a method for automatic evaluation of machine translation.
\newblock In \emph{Proceedings of the 40th annual meeting of the Association
  for Computational Linguistics}, pages 311--318.

\bibitem[{Qiang et~al.(2020)Qiang, Li, Zhu, Yuan, and Wu}]{qiang2020lexical}
Jipeng Qiang, Yun Li, Yi~Zhu, Yunhao Yuan, and Xindong Wu. 2020.
\newblock Lexical simplification with pretrained encoders.
\newblock In \emph{Proceedings of the AAAI Conference on Artificial
  Intelligence}, volume~34, pages 8649--8656.

\bibitem[{Rennie et~al.(2017)Rennie, Marcheret, Mroueh, Ross, and
  Goel}]{rennie2017self}
Steven~J Rennie, Etienne Marcheret, Youssef Mroueh, Jerret Ross, and Vaibhava
  Goel. 2017.
\newblock Self-critical sequence training for image captioning.
\newblock In \emph{Proceedings of the IEEE Conference on Computer Vision and
  Pattern Recognition}, pages 7008--7024.

\bibitem[{Salazar et~al.(2020)Salazar, Liang, Nguyen, and
  Kirchhoff}]{salazar2020masked}
Julian Salazar, Davis Liang, Toan~Q Nguyen, and Katrin Kirchhoff. 2020.
\newblock Masked language model scoring.
\newblock In \emph{Proceedings of the 58th Annual Meeting of the Association
  for Computational Linguistics}, pages 2699--2712.

\bibitem[{Smith et~al.(2016)Smith, Turner, Sanford-Moore, and
  Koons}]{Smith2016TheLF}
Malbert Smith, J.~Turner, Eleanor~E. Sanford-Moore, and Heather~H. Koons. 2016.
\newblock The lexile framework for reading: An introduction to what it is and
  how to use it.

\bibitem[{Speer et~al.(2018)Speer, Chin, Lin, Jewett, and
  Nathan}]{robyn_speer_2018_1443582}
Robyn Speer, Joshua Chin, Andrew Lin, Sara Jewett, and Lance Nathan. 2018.
\newblock \href {https://doi.org/10.5281/zenodo.1443582} {Luminosoinsight /
  wordfreq: v2.2}.
\newblock Doi.org/10.5281/zenodo.1443582.

\bibitem[{Stahlberg and Kumar(2020)}]{stahlberg2020seq2edits}
Felix Stahlberg and Shankar Kumar. 2020.
\newblock Seq2edits: Sequence transduction using span-level edit operations.
\newblock In \emph{Proceedings of the 2020 Conference on Empirical Methods in
  Natural Language Processing (EMNLP)}, pages 5147--5159.

\bibitem[{Surya et~al.(2019)Surya, Mishra, Laha, Jain, and
  Sankaranarayanan}]{surya2019unsupervised}
Sai Surya, Abhijit Mishra, Anirban Laha, Parag Jain, and Karthik
  Sankaranarayanan. 2019.
\newblock Unsupervised neural text simplification.
\newblock In \emph{Proceedings of the 57th Annual Meeting of the Association
  for Computational Linguistics}, pages 2058--2068.

\bibitem[{Thomas and Anderson(2012)}]{thomas2012wordnet}
S~Rebecca Thomas and Sven Anderson. 2012.
\newblock Wordnet-based lexical simplification of a document.
\newblock In \emph{KONVENS}, pages 80--88.

\bibitem[{Trischler et~al.(2017)Trischler, Wang, Yuan, Harris, Sordoni,
  Bachman, and Suleman}]{trischler2017newsqa}
Adam Trischler, Tong Wang, Xingdi Yuan, Justin Harris, Alessandro Sordoni,
  Philip Bachman, and Kaheer Suleman. 2017.
\newblock Newsqa: A machine comprehension dataset.
\newblock In \emph{Proceedings of the 2nd Workshop on Representation Learning
  for NLP}, pages 191--200.

\bibitem[{Wang et~al.(2020)Wang, Cho, and Lewis}]{wang2020asking}
Alex Wang, Kyunghyun Cho, and Mike Lewis. 2020.
\newblock Asking and answering questions to evaluate the factual consistency of
  summaries.
\newblock In \emph{Proceedings of the 58th Annual Meeting of the Association
  for Computational Linguistics}, pages 5008--5020.

\bibitem[{Wubben et~al.(2012)Wubben, van~den Bosch, and
  Krahmer}]{wubben2012sentence}
Sander Wubben, Antal van~den Bosch, and Emiel Krahmer. 2012.
\newblock Sentence simplification by monolingual machine translation.
\newblock In \emph{Proceedings of the 50th Annual Meeting of the Association
  for Computational Linguistics (Volume 1: Long Papers)}, pages 1015--1024.

\bibitem[{Xu et~al.(2015)Xu, Callison-Burch, and Napoles}]{Xu2015ProblemsIC}
W.~Xu, Chris Callison-Burch, and Courtney Napoles. 2015.
\newblock Problems in current text simplification research: New data can help.
\newblock \emph{Transactions of the Association for Computational Linguistics},
  3:283--297.

\bibitem[{Xu et~al.(2016)Xu, Napoles, Pavlick, Chen, and
  Callison-Burch}]{xu2016optimizing}
Wei Xu, Courtney Napoles, Ellie Pavlick, Quanze Chen, and Chris Callison-Burch.
  2016.
\newblock Optimizing statistical machine translation for text simplification.
\newblock \emph{Transactions of the Association for Computational Linguistics},
  4:401--415.

\bibitem[{Zhang and Bansal(2019)}]{Zhang2019AddressingSD}
Shiyue Zhang and Mohit Bansal. 2019.
\newblock Addressing semantic drift in question generation for semi-supervised
  question answering.
\newblock In \emph{Proceedings of the 2019 Conference on Empirical Methods in
  Natural Language Processing and the 9th International Joint Conference on
  Natural Language Processing (EMNLP-IJCNLP)}.

\bibitem[{Zhang and Lapata(2017)}]{zhang2017sentence}
Xingxing Zhang and Mirella Lapata. 2017.
\newblock Sentence simplification with deep reinforcement learning.
\newblock In \emph{Proceedings of the 2017 Conference on Empirical Methods in
  Natural Language Processing}, pages 584--594.

\bibitem[{Zhang et~al.(2020)Zhang, Merck, Tsai, Manning, and
  Langlotz}]{zhang2020optimizing}
Yuhao Zhang, Derek Merck, Emily Tsai, Christopher~D Manning, and Curtis
  Langlotz. 2020.
\newblock Optimizing the factual correctness of a summary: A study of
  summarizing radiology reports.
\newblock In \emph{Proceedings of the 58th Annual Meeting of the Association
  for Computational Linguistics}, pages 5108--5120.

\bibitem[{Zhong et~al.(2020)Zhong, Jiang, Xu, and Li}]{zhong2020discourse}
Yang Zhong, Chao Jiang, Wei Xu, and Junyi~Jessy Li. 2020.
\newblock Discourse level factors for sentence deletion in text simplification.
\newblock In \emph{Proceedings of the AAAI Conference on Artificial
  Intelligence}, volume~34, pages 9709--9716.

\bibitem[{Zhu et~al.(2010)Zhu, Bernhard, and Gurevych}]{zhu2010monolingual}
Zhemin Zhu, Delphine Bernhard, and Iryna Gurevych. 2010.
\newblock A monolingual tree-based translation model for sentence
  simplification.
\newblock In \emph{Proceedings of the 23rd International Conference on
  Computational Linguistics (Coling 2010)}, pages 1353--1361.

\end{thebibliography}
\bibliographystyle{acl_natbib}

\clearpage
\appendix
\renewcommand{\thetable}{A\arabic{table}}
\setcounter{table}{0}
\renewcommand{\thefigure}{A\arabic{figure}}
\setcounter{figure}{0}

\section*{Ethical Considerations}

We present a method for text simplification and verify its performance on text from the news domain in the English language. Even though we expect the method to be adaptable to other domains and languages, we have not verified this assumption experimentally and limit our claims to the English news domain.

When comparing to prior work (e.g., ACCESS model), we obtained implementations directly from the authors (through Github repositories) and produced results following the recommended setting, with an objective to present prior work as a strong comparison point.

For the human evaluation, we paid the annotators above the minimum wage, and did not collect any personal identifiable information. We selected  topics to avoid  sensitive or political subjects and had our protocols reviewed by the university's IRB committee (Protocol ID: 2018-07-11230). We relied on a third party (Amazon Mechanical Turk) to remunerate the crowd-workers.

\section{Appendices}
\label{sec:appendix}

\subsection{Training Details}
\label{appendix:training_details}
We detail the model architecture size, data, optimizer of the models we train in the paper. All models were trained using Pytorch and HuggingFace's Transformers library\footnote{https://github.com/huggingface/transformers}. We use the Apex\footnote{https://github.com/nvidia/apex} library to enable half-precision training.

The KiS procedure was trained on a single GPU, either an Nvidia V-100 (16Gb memory) or a Quadro RTX 8000 (48 Gb memory). We ran a total of around 200 experiments, with an average run-time of one week.

Because the procedure is unsupervised, the model was trained using a large unreleased corpus of news articles, containing 7 million news articles in English. 

\textbf{KiS Model} is initialized with a \textit{GPT2-medium} model. We used the Adam optimizer, with a learning rate of $10^{-6}$, a batch-size of 1, using $k$-SCST with $k=8$. 

\textbf{Finetune Baseline} is initialized with a \textit{GPT2-medium} model. We train using  using standard teacher forcing on the 40,000 samples in the \textit{paired Newsela dataset}, reserving 2,000 samples for validation. We use the Adam optimizer, and use the validation set to choose a learning rate of $10^{-5}$, and a batch-size of 8, and run for 3 epochs before seeing a plateau in the validation loss.

\textbf{Discriminator Model} is initialized with a \textbf{Roberta-base}, and retrained every time the training buffer reaches 2,000 samples. The discriminator is reset to the original \textit{Roberta-base} each time the training buffer is full. We use a standard cross-entropy loss, the ADAM optimizer with a learning rate of $10^{-5}$ and a batch size of 8. Each time we retrain, we run for 5 epochs, and checkpoint one model after each epoch. The checkpoint that achieves the highest performance on a validation set becomes the new discriminator for the next round.

\subsection{Human Evaluation Instructions}
Figure~\ref{fig:user_instructions} shows the instructions given to crowd-worker participants for the manual evaluation. 

\label{appendix:user_instructions}
\begin{figure}[!htbp]
    \begin{framed}
• The entire HIT should take no more than 15 minutes:\\
    (1) You will answer a pre-questionnaire.\\
    (2) Read 4 short news stories and answer comprehension questions about each.\\
• If you believe the answer is not in the document, you can select the option ``Answer not in document''.\\
• There is no time limit for each individual document or question.\\
• You can leave at any point but will not complete the HIT.\\
• You can complete this task at most once.\\
• If you have a question/problem, contact us at \textit{email}.\\
\end{framed}
\caption{Instructions given to participants of the comprehension evaluation. Participants were recruited on Amazon Mechanical Turk (MTurk), on which jobs are named ``HIT''.}
\label{fig:user_instructions}
\end{figure}

\subsection{Full Example of Generated Texts}
\begin{figure*}[!htbp]
    \includegraphics[width=\textwidth]{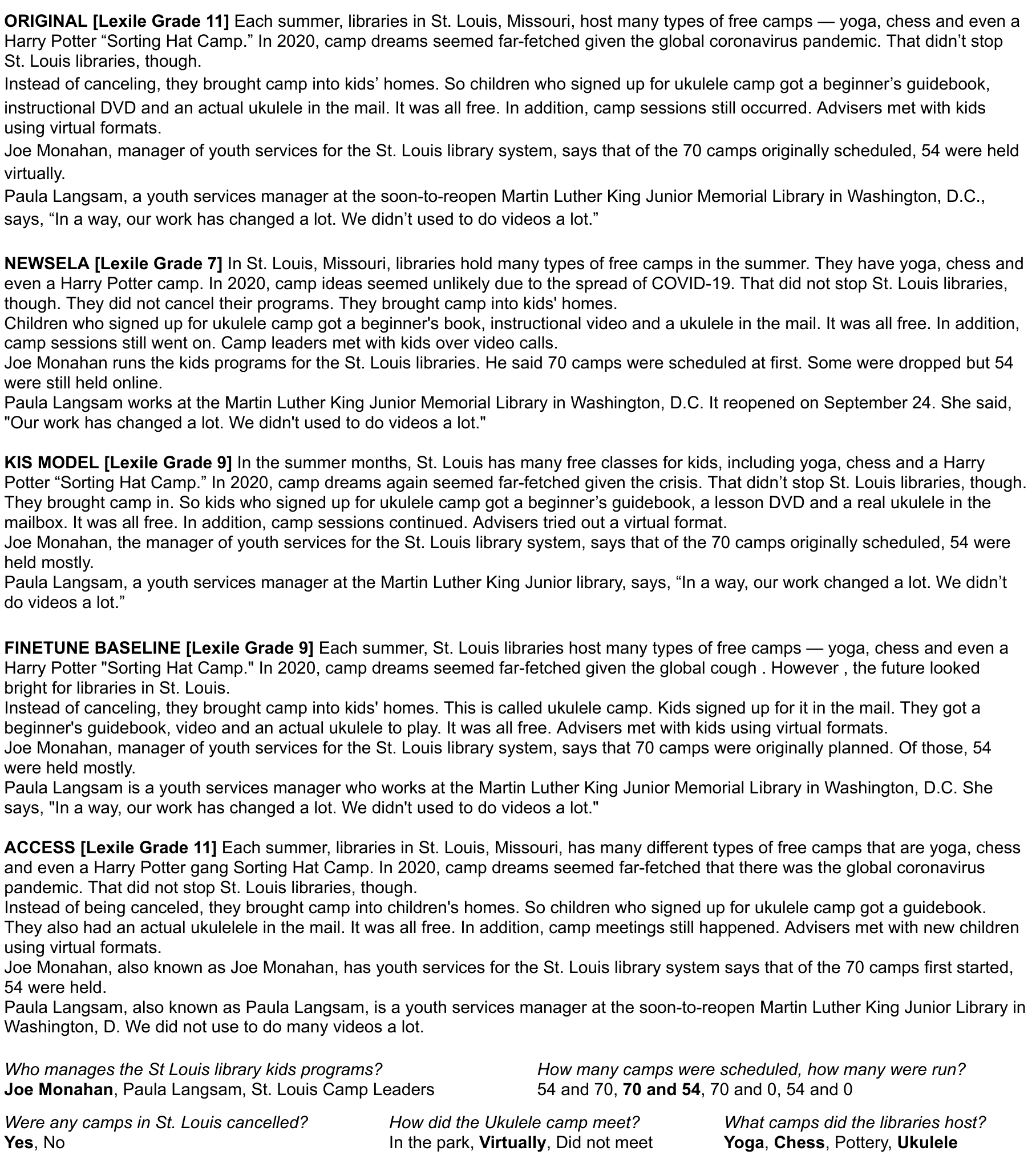}

    \caption{\textbf{Complement to Figure~\ref{fig:user_study_example}.} Example Task for the Comprehension Study. Participants were assigned to one of five settings: original, Newsela, KiS, Finetune Baseline, and ACCESS. Participants were instructed to answer the five comprehension questions.}
    \label{fig:complement_study_examples}
\end{figure*}

Figure~\ref{fig:complement_study_examples} is a complement to Figure~\ref{fig:user_study_example}, with the five stimuli that were shown for the \textit{Covid Libraries} document.

\subsection{Detailed of Human Evaluation Results}

Table~\ref{table:full_study_table} details the timing and number of participants for each combination of document and stimuli.

\label{appendix:full_study_table}
\begin{table*}[!htbp]
    \centering
    \resizebox{0.7\textwidth}{!}{%
    \begin{tabular}{llllll}
    & \multicolumn{5}{c}{\textbf{Simplification Model}} \\ \hline
\multicolumn{1}{l|}{\textbf{Document Id}}       & \textbf{Original} & \textbf{Newsela} & \textbf{Sup. Base.} & \textbf{ACCESS} & \multicolumn{1}{l|}{\textbf{KiS}} \\ \hline
\multicolumn{1}{l|}{Marvel Show}                & 152 (12)          & 209 (11)         & 140 (11)            & 209 (14)        & \multicolumn{1}{l|}{126 (13)}     \\
\multicolumn{1}{l|}{Covid Libraries}            & 167 (14)          & 180 (12)         & 182 (10)            & 190 (13)        & \multicolumn{1}{l|}{171 (12)}     \\
\multicolumn{1}{l|}{Sustainable Food}           & 163 (13)          & 144 (10)         & 181 (13)            & 242 (13)        & \multicolumn{1}{l|}{154 (12)}     \\
\multicolumn{1}{l|}{Iceberg Collision}          & 208 (14)          & 116 (11)         & 139 (12)            & 104 (12)        & \multicolumn{1}{l|}{119 (12)}     \\ \hline
\multicolumn{1}{l|}{\textbf{Version Aggregate}} & 174 (53)          & 163 (44)         & 161 (46)            & 188 (52)        & \multicolumn{1}{l|}{143 (49)}     \\ \hline
\end{tabular}
    }
    \caption{\textbf{Average time taken and number of participants in each of the  document/stimuli combinations.} Also shown are  aggregates (mean time taken and total number of participants).}
    \label{table:full_study_table}
\end{table*}

\begin{table*}
    \centering
    \begin{tabular}{lllrrll}
    \hline
    Model             & SARI  & BLEU  & \%FKGL & \%Lexile & Comp. & Cov.  \\ \hline
    KiS Full          & 0.709 & 0.526 & 100    & 72       & 0.85  & 0.636 \\ \hline
    KiS No Fluency    & 0.718 & 0.611 & 99     & 95       & 1.02  & 0.901 \\
    KiS No Salience   & 0.695 & 0.591 & 100    & 65       & 1.01  & 0.701 \\
    KiS No Simplicity & 0.672 & 0.617 & 51     & 23       & 0.92  & 0.809 \\ \hline
    \end{tabular}
    \caption{\textbf{Automatic results of the three ablation models.} \textit{SARI} and \textit{BLEU} are reference-based metrics. \textit{\% FKGL} and \textit{\% Lexile} are the percentage of simplified paragraphs with a lower FKGL and Lexile score than the original paragraph. \textit{Comp.} is the average compression ratio (\# of words), and \textit{Cov.} is the average coverage score of the simplifications.}
    \label{table:ablation_study}
\end{table*}
\subsection{Detail of Ablation Study Results}
\label{appendix:ablation_study}
Table~\ref{table:ablation_study} details the metric results of the three ablated models, an extension to Table~\ref{tab:automatic_results}. An example output of each ablated model, illustrating the limitation when a score component is missing, is given in Figure~\ref{fig:ablation_examples}.

One surprising element is that the model trained without fluency achieves higher scores on almost all metrics, compared to the full model. This surprising fact is due to the fact that without fluency, the model does not learn to generate full sentences (see the example in Figure~\ref{fig:ablation_examples}). Instead, the model learns to concatenate high-scoring phrases together, which can boost automatic metrics artificially. In fact, the strong performance of a model generating incomplete sentences reveals a limitation of current automatic metrics, such as BLEU and SARI.

\end{document}